\pdfoutput=1

\documentclass[11pt]{article}

\usepackage[final]{acl}

\usepackage{times}
\usepackage{latexsym}

\usepackage[T1]{fontenc}

\usepackage[utf8]{inputenc}

\usepackage{microtype}

\usepackage{inconsolata}

\usepackage{amsmath}
\usepackage{amsfonts,amssymb}
\usepackage{tikz}
\usepackage{float}
\usepackage{subcaption} 
\usepackage{multirow}
\usepackage{booktabs}
\usepackage{tabularx}
\usepackage[normalem]{ulem}
\useunder{\uline}{\ul}{}
\title{Align-to-Distill:  Trainable Attention Alignment for \\ Knowledge Distillation in Neural Machine Translation}

\author{Heegon Jin, Seonil Son, Jemin Park, Youngseok Kim, Hyungjong Noh, Yeonsoo Lee\\
        NC Research, AI Tech Center \\
        Seongnam, South Korea \\
         \{heegon, deftson, hbk5844, kys159, nohhj0209, yeonsoo\}@ncsoft.com\\}

\begin{document}
\maketitle
\begin{abstract}
The advent of scalable deep models and large datasets has improved the performance of Neural Machine Translation~(NMT). 
Knowledge Distillation (KD) enhances efficiency by transferring knowledge from a teacher model to a more compact student model.
However, KD approaches to Transformer architecture often rely on heuristics, particularly when deciding which teacher layers to distill from. 
In this paper, we introduce the ``Align-to-Distill'' (A2D) strategy, designed to address the feature mapping problem by adaptively aligning student attention heads with their teacher counterparts during training.
The Attention Alignment Module~(AAM) in A2D performs a dense head-by-head comparison between student and teacher attention heads across layers, turning the combinatorial mapping heuristics into a learning problem. 
Our experiments show the efficacy of A2D, demonstrating gains of up to +3.61 and +0.63 BLEU points for WMT-2022 De→Dsb and WMT-2014 En→De, respectively, compared to Transformer baselines. \footnote{The code and data are available at \url{https://github.com/ncsoft/Align-to-Distill}.}
\end{abstract}
\section{Introduction}
Transformer-based encoder-decoder models have achieved remarkable success in various natural language processing tasks \cite{Vaswani2017, Devlin2019, Liu2019, Lewis2020}, including Neural Machine Translation (NMT) \cite{Sutskever2014, Bahdanau2015, Sennrich2016, Vaswani2017}. 
However, the autoregressive decoding process often imposes a significant computational burden, especially as the number of layers and parameters escalates with increasing model complexity.
This presents substantial challenges when deploying the Transformer-based models for real-time applications~\cite{gu2017} and online services~\cite{zhou2022}.

One of the possible solutions is to reduce the model size using knowledge distillation (KD) \cite{Hinton2015}. 
KD facilitates the transfer of knowledge from a high-performing, large-parameter teacher model to a more moderately sized student model. 
This process alleviates deployment challenges by generating a distilled model that is both lightweight and efficient, ensuring reduced inference times and lower computational resource requirements.
Furthermore, with the guidance of the teacher model, the student model can potentially achieve performance levels closer to those of the teacher model compared to training it without the teacher's assistance.

KD, initially proposed by \citeauthor{Bucila2006, Ba2014, Hinton2015}, involves transferring knowledge to the student model using responses from the network's last layer. 
Among its variants, Sequence-level KD~\cite{kim2016} and Selective KD~\cite{Wang2021a} leverage the final output and soft labels from teacher's responses, respectively. 
These strategies can be categorized as response-based KD~\cite{DBLP:journals/ijcv/GouYMT21}.
Meanwhile, not only the final layer outputs are used, but intermediate features from the teacher model's layer are also used as a medium for a more effective and comprehensive distillation of knowledge~\cite{Romero2015, Zagoruyko2017, Sun2019, Jiao2020, Sun2020}. These approaches belong to the category of feature-based KD~\cite{DBLP:journals/ijcv/GouYMT21}.
Most feature-based KD in Transformers has concentrated on compressing encoder-based models~\cite{Sanh2019,Sun2019,Jiao2020,Wang2020,Sun2020,Passban2021}, including pre-trained models like BERT~\cite{Devlin2019}.
On the other hand, some studies~\cite{Wu2020, Shleifer2020PreSum} have applied feature-based KD to the decoder for generative tasks.
However, they found it less effective compared to response-based KD for decoder distillation~\cite{kim2016,Kasai2020DeepES,Wang2021a}.

While extending KD to features across the layers does enrich knowledge transfer, it prompts an open question: `From which teacher layer should the student layer learn and from which should it not?'. 
Instead of resolving this issue through trainable methods, several studies~\cite{Sun2019, Jiao2020, Wu2020, Passban2021} have circumvented the issue using heuristic approaches.
Those approaches require a heuristic skip or combination of teacher layers to align with the student layer.
However, as the number of layers increases, the complexity of heuristically selecting features grows, necessitating an exhaustive search for the optimal combination strategy.
For example, Combinatorial KD~\cite{Wu2020} demonstrated that its peak performance relies on language-pair-specific feature mapping. 

In this paper, we introduce a novel KD strategy, Align-to-Distill (A2D), that addresses the feature mapping problem using a trainable Attention Alignment Module (AAM). 
Unlike earlier KD methods that relied on combinatorial feature mapping heuristics, A2D provides an end-to-end trainable solution.
The adaptive alignment of features removes the necessity for a data-dependent mapping strategy.
Furthermore, AAM aligns the student attention map in each head with those of the teacher, resulting in more effective distillation compared to layer-wise feature mapping. 
AAM enables each attention head in the student model to be compared with every head in the teacher model across different layers, by employing pointwise convolution with only a few additional parameters. As a result, there is no longer a need for head or layer parity between the student and teacher models.

Notably, our experimental results and analysis show that due to its fine-grained attention transfer in a head-wise manner, A2D is effectively applicable to the decoder of the Transformer, an area where previous feature-based KD approaches have typically struggled.
By compressing the decoder with A2D, we could reduce the cost of autoregressive inference while preserving its performance.
Our comprehensive studies on both high-resource and low-resource NMT tasks show that our method consistently outperforms state-of-the-art baselines, spanning both feature-based and response-based KD methods.
In particular, even with a smaller model size than the teacher, students trained with A2D can match or even surpass teacher performance in low-resource settings. 

Our contribution is three-folded as follows:

\begin{itemize}
\item We introduce ``Align-to-Distill'' (A2D), a novel attention-based distillation method that can be effectively applied to the decoder of the transformer.
\item A2D overcomes the limitations imposed by feature-mapping heuristics of previous distillation approaches by introducing a learnable alignment between attention heads across different layers. 
\item A2D enables fine-grained attention knowledge transfer from teacher to student, thereby outperforming state-of-the-art KD strategies in both high-resource and low-resource translation tasks.
\end{itemize}
\section{Preliminaries} 
\subsection{Multi-Head Attention}
\label{section:2.1}
\noindent\textbf{Multi-head Attention (MHA)} is a core component of the Transformer architecture introduced by \citeauthor{Vaswani2017}. We use $\mathbf{X}\in\mathcal{R}^{L\times d}$ to denote the input to the Transformer layer, which is a sequence of embeddings fed into MHA, where $L$ is the length of the text input and $d$ is the dimension of model embedding.

In the Transformer layer, the input sequence $\mathbf X$ is mapped to three unique vector embeddings: the query ($\mathbf{Q}$), the key ($\mathbf{K}$), and the value ($\mathbf{V}$). Each of these embeddings is associated with its own projection weight, denoted by $\mathbf{W}_q$, $\mathbf{W}_k$, and $\mathbf{W}_v$ respectively.
\begin{equation}
    \mathbf{Q}=\mathbf{X}\mathbf{W}_q, \mathbf{K}=\mathbf{X}\mathbf{W}_k, \mathbf{V}=\mathbf{X}\mathbf{W}_v
\end{equation}

Note that these projection weights ($\mathbf{W}_{q|k|v}$) are unique for each attention head and layer. 
The shape of $\mathbf{W}_{q|k|v}$ is $\mathcal{R}^{{d} \times d_{\mathrm{head}}}$, which results in $\mathbf{Q,~K,~V} \in \mathcal{R}^{{L}\times d_{\mathrm{head}}}$. Head dimension $d_\mathrm{head}$ is $d$ divided by the number of attention heads in a layer.

Scaled dot-product attention, $\mathrm{attn}(\mathbf{X})$, that each MHA head computes is defined as follows:
\begin{equation}
\small
\label{eq:sdp}
\mathrm{attn}(\mathbf{X}) =
s\Big(\mathbf{Q}\mathbf{K}^{\top} / \sqrt{d_{\mathrm{head}}}\Big)\mathbf{V} = \mathbf{H}\mathbf{V} \in\mathcal{R}^{L\times d_{\mathrm{head}}}
\end{equation}
where $s(\cdot)$ denotes a softmax function. 
The attention output denoted by $\mathrm{attn(\mathbf{X})}$ is concatenated over $d_\mathrm{head}$ to form MHA of shape $\mathcal{R}^{L \times d}$.

\noindent\textbf{Attention map} denoted by $\mathbf{H}$ above, is the scaled dot-product of the query-key for each head, represented as $s(\mathbf{QK^\top}/\sqrt{d_{\mathrm{head}})} \in \mathcal{R}^{L \times L}$.
To make clearer notion of distinct heads and layers, throughout the paper the attention map will appear as:

\begin{equation}
\small
    \mathbf{H}_{(m,n)} = s\left(\mathbf{Q}_{(m,n)}\mathbf{K}^\top_{(m,n)}/\sqrt{d_\mathrm{head}}\right)\in\mathcal{R}^{L\times L}
\label{eq:head}
\end{equation}
where $m$ and $n$ serve as indices for head and layer inside the whole Transformer architecture. 
Thus, $\mathbf{H}_{(m,n)}$ represents as attention map $m$-th head in $n$-th Transformer layer.
Similarly $\mathbf{Q}_{(m,n)}$, $\mathbf{K}_{(m,n)}$, and $\mathbf{V}_{(m,n)}$ refer to the query, key, and value of the same head.

\subsection{Knowledge Distillation}
The idea of knowledge distillation from a larger neural network to a smaller network was proposed in \cite{Hinton2015}. In the paper, the output distribution of the larger teacher network is used as a soft target for training the student network.
\begin{equation}
\mathcal{L}_\mathrm{KD} = -\sum_{x \in \mathcal{X}}p^T\mathrm{log}(p^S)
\label{eq:hintonkd}
\end{equation}
where $p^T$ and $p^S$ denote the output distributions of the teacher and student.
The core idea of training the student to imitate the teacher extends further from comparing the output distribution to any valuable intermediate features. In general, the following loss function provides an abstraction.

\begin{equation}
\mathcal{L} = \sum_{x \in \mathcal{X}}\mathrm{D}(f^T(x),f^S(x))
\label{eq:kdreview}
\end{equation}
Equation~\eqref{eq:kdreview} compares the features of the teacher and student models ($f^T(x)$, $f^S(x)$)  for a given input $x$ from a dataset $\mathcal X$ with a measure of dissimilarity (D). 
There are a number of widely used choices for features to distill knowledge from, such as hidden states \cite{Sun2019, Passban2021} or attention maps \cite{Jiao2020, Wu2020}.
A common choice for $\mathrm{D}$ is Kullback-Leibler Divergence~\cite{KLD} or mean-squared error. 
As the student network learns to imitate the features from the teacher with Equation~\eqref{eq:kdreview},  the teacher model's knowledge, represented by the feature $f^T(x)$, is transferred to the student. 
This transfer of knowledge helps the student model converge to a better optimum that it could not reach on its own. 
\section{Related Works}
\begin{figure*}[t]
\centering
  \includegraphics[width=1.0\linewidth]{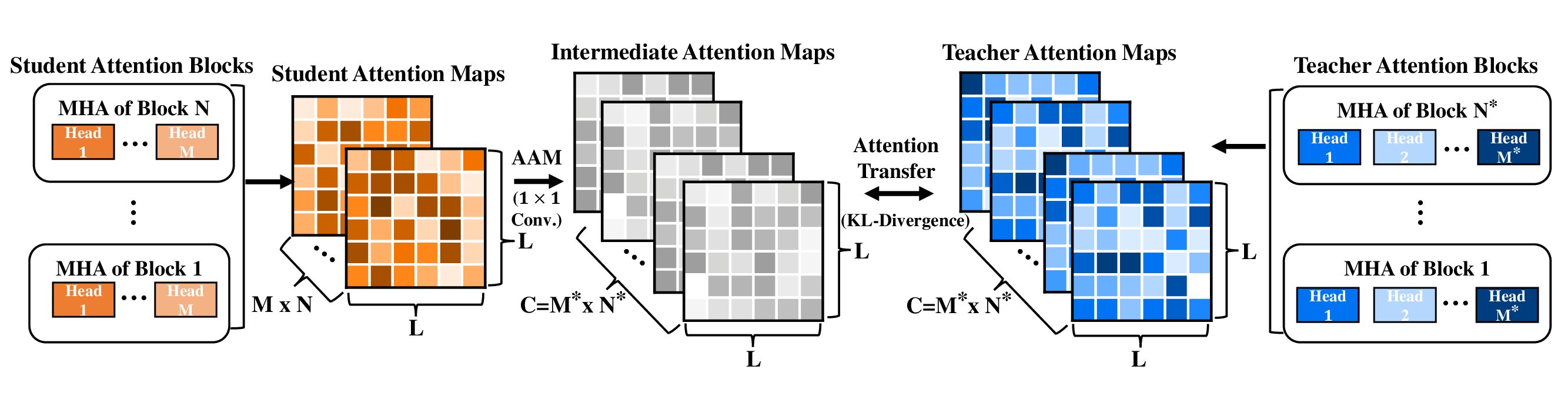}
\caption{
Attention Transfer with A2D. 
The Attention Alignment Module~(AAM), implemented as a pointwise convolution layer, produces intermediate attention maps from a collection of student attention maps.
The number of intermediate maps matches the total attention maps of the teacher model, encompassing all layers and heads.
These intermediate maps are then directly compared to the teacher's attention maps using KL-Divergence, without any form of reduction.}
  \label{fig:framework}
\end{figure*}
In the field of feature-based KD studies, which incorporate intermediate representations of the teacher for KD, an aligning strategy has been explored to handle different architectural settings between the teacher and student. Since the teacher model generally has more layers than the student model, Patient KD \cite{Sun2019} proposed skipping every other layer or using only the last consecutive layers of the teacher model so that the number of matching layers becomes equal. Likewise, TinyBERT \cite{Jiao2020} formulated a layer-mapping function that selects hint layers to match the student layers at a similar depth level. Meanwhile, MiniLM \cite{Wang2020} circumvented the aligning issue by distilling only the last layer. However, it constrained the number of heads to be equivalent between teacher and student networks.

To address the issue of skipped teacher layers, CKD \cite{Wu2020} proposed layer fusion, a method that projects several teacher layers into one fused layer. The fusion is then matched with the corresponding student layer, enabling distillation from all teacher layers. Although CKD benefits from the representations contained in more teacher layers, the mapping between fused teacher layers and student layers is set heuristically, before training. As a result, its optimal combination varies due to the training dataset, and the number of cases for mapping becomes intractable when the number of layers increases. Additionally, performance degradation was reported when the method was applied to the decoder part of the Transformer, regardless of the combination setting. This led to CKD being applied solely to the encoder side of the Transformer. While ALP-KD \cite{Passban2021} made the mapping partially adaptive with an attention mechanism, it still requires dividing student layers into buckets, which involves heuristics, and the result is dependent on how the buckets are grouped. Both CKD and ALP-KD are layer-wise fusion distillation methods and do not consider more detailed mappings which can be found in attention heads. 
\section{Methodology}
In this section, we provide a detailed overview of the architecture and training of the Attention Alignment Module (AAM), the core component of our Align-to-Distill (A2D) approach. 
Figure~\ref{fig:framework} illustrates the overall framework.

\subsection{Attention Map as a Knowledge}
The attention mechanism is a vital component in Transformer models, capturing the context and relationships between words in a sentence.
Attention alignment ensures that the student model, despite potential differences in network structure compared to the teacher model, focuses on the same word relationships and contextual nuances as the teacher.
This process allows the student model to gain insights beyond mere output mimicry; it learns the underlying relationships that the teacher model uses for prediction, effectively transferring knowledge from the teacher model to the student.

A2D compares the attention maps in Equation~\eqref{eq:head} from the teacher and student for knowledge distillation.
As depicted previously the shape of the attention map is determined by the length of the input (i.e. $\mathbf{H}_{(m,n)}\in\mathcal{R}^{L\times L}$), not by the model hyperparameters.
This eases the comparison between knowledge features between the teacher and student models on a different scale.

In the encoder-decoder model, there exist three distinct attention that computes relation within, and in between the encoder and decoder input sequences: self-attention within the encoder and decoder, and cross-attention from encoder to decoder.
The A2D method incorporates all three types of attention maps for effective knowledge distillation:
\begin{align}
    &\mathbf{H}_{(m,n)}^\mathrm{enc-self}\in\mathcal{R}^{L_{\mathrm{src}}\times L_{\mathrm{src}}},\label{eq:parts1} \\
    &\mathbf{H}_{(m,n)}^\mathrm{dec-self}\in\mathcal{R}^{L_{\mathrm{tgt}}\times L_{\mathrm{tgt}}},\label{eq:parts2} \\
    &\mathbf{H}_{(m,n)}^\mathrm{dec-cross}\in\mathcal{R}^{L_{\mathrm{src}}\times L_{\mathrm{tgt}}}\label{eq:parts3}
\end{align}
Note that $L_\mathrm{src}$ and $L_\mathrm{tgt}$ are the lengths of the source and target sentences for machine translation, which make the shape of attention maps among three types of attention.
During the attention knowledge transfer, each type of attention map is compared between teacher and student, respectively.

\subsection{Attention Alignment Module}
\subsubsection{Module Architecture}\label{section:AAM}
The student and teacher models share the same $L_{\mathrm{src}}$ and $L_{\mathrm{tgt}}$ during distillation, leading to attention maps of the same shape for both models, regardless of model differences. 
However, the number of attention maps in these models differs since the teacher model has more layers and more heads in each layer compared to the student model.
This discrepancy in numbers makes it infeasible to make a straightforward one-to-one mapping of attention maps between teacher and student models. 
To resolve this, we devise an Attention Alignment Module~(AAM) that generates intermediate attention maps, bridging the gap between the two groups of attention maps.

AAM performs pointwise convolution ($1\times 1$ Conv. as in Figure~\ref{fig:framework}) on the student attention maps. 
This operation creates an equal number of intermediate attention maps to the teacher maps. 
Each intermediate map results from a weighted sum of all student attention maps. 
Consequently, a one-to-one comparison between the teacher maps and the intermediate maps simulates transferring knowledge with a fully-connected mapping between teacher and student attention maps.

Two desirable attributes of pointwise convolution take major roles in effective attention knowledge transfer: (1) convolution operation preserves sequential information in each attention map, while (2) allowing fully-connected weighted mapping between the groups. 

We can represent the intermediate attention maps, $\mathbf{H}^I_c$, from student attention maps, $\mathbf{H}^S_{(m,n)}$ with the learnable parameters of AAM:
\begin{equation}
\label{eq:intattmap}
    \mathbf{H}^I_{c} = \sum_{m=1}^{M}\sum_{n=1}^{N} w_{(m,n),c} \mathbf{H}^S_{(m,n)} + b_{c},  
    \\ (c\in{1,...,C})
\end{equation}

\noindent Here, $w_{(m,n),c}$ and $b_c$ represent the weight and bias of AAM, respectively. 
These parameters generate the $c$-th intermediate attention map ($c\in{1,..., C}$), where $C$ denotes the total number of teacher's attention heads. 
The variables $M$ and $N$ denote the number of student attention heads within a single layer, and the maximum layer depth of the student respectively. 

In summary, AAM is a pointwise convolution layer that generates $C$ intermediate attention maps from a total of $M*N$ student attention maps. 
The convolution operation of AAM ensures that a one-to-one comparison of intermediate maps to teacher attention maps functions similarly to a fully-connected comparison between student and teacher attention maps. 
The trained weights of AAM ($w_{(m,n),c}$) after the distillation can be used to analyze the alignment between student and teacher attention heads (discussed further in the Analysis section; see Figure~\ref{fig:analysis}).

Finally, the AAM only adds a small number of extra parameters and operations when determining map discrepancy, compared to the student model. These are estimated as $M*N*C$ for parameters and $C*L^2$ for operations. After training, the AAM can be discarded, as it serves no function during inference.

\subsubsection{Module Training}
AAM generates intermediate attention maps (denoted as $\mathbf{H}^I_c$) that have the same number as the teacher attention maps, referred to as $\mathbf{H}^T_c$.
We minimize the KL-Divergence between the attention distributions of the intermediate attention maps and the teacher attention maps as:
\begin{equation}\label{eq:regatt}
    \mathcal{L}_{att} = \sum_{c=1}^{C} D_{KL}({\mathbf{H}^T_c} || {\mathbf{H}^I_c}).
\end{equation}
This equation represents the attention transfer loss function, $\mathcal{L}_\mathrm{att}$, which quantifies the difference between the teacher's attention maps and the intermediate attention maps. 
To break down further, $\mathcal{L}_\mathrm{att}$ incorporates three different terms according to the types of attention maps.

\begin{equation}
\small
\mathcal{L}_\mathrm{att} = \mathcal{L}_\mathrm{att}^{\mathrm{enc-self}} + 
\frac{1}{2}\left(\mathcal{L}_\mathrm{att}^\mathrm{dec-self} +
\mathcal{L}_\mathrm{att}^{\mathrm{dec-cross}}\right)
\label{eq:attnloss}
\end{equation}
$\mathcal{L}_\mathrm{att}^{\mathrm{enc-self}}$, $\mathcal{L}_\mathrm{att}^{\mathrm{dec-self}}$, and $\mathcal{L}_\mathrm{att}^{\mathrm{dec-cross}}$ above correspond to the loss in Equation~\eqref{eq:regatt} estimates from three different types of attention maps in Equation~\eqref{eq:parts1}, \eqref{eq:parts2}, \eqref{eq:parts3}. 
To balance the loss scale between the encoder and decoder, we divide the sum of losses for the decoder by 2. 
To sum up, the attention transfer loss $\mathcal{L}_\mathrm{att}$ incorporates three types of attention mechanism in the encoder-decoder model (i.e. self-attention of each encoder and decoder, and cross attention)
with balancing for the encoder and decoder-side losses.

Since $\mathcal{L}_\mathrm{att}$  is orthogonal to the loss from knowledge distillation by \citeauthor{Hinton2015} ($\mathcal{L}_\mathrm{KD}$), it can be used jointly to give additional supervision.  
Throughout the paper, we denote the distillation approach with only $\mathcal{L}_\mathrm{KD}$ as vanilla KD. 
To form a final loss for A2D, we add the cross-entropy loss $\mathcal{L}_\mathrm{CE}$ of student model for translation task as follows:
\begin{equation}\label{eq:ceatt}
    \mathcal{L} = \mathcal{L}_\mathrm{CE} + \lambda\mathcal{L}_\mathrm{att} + \mu\mathcal{L}_\mathrm{KD}
\end{equation}
The student model and the AAM are collaboratively optimized through an end-to-end approach, as described in Equation~\eqref{eq:ceatt}. 
Given that $H_c^I$ is differentiable with respect to $w_c$, the $\mathcal{L}_\mathrm{att}$ from Equation~\eqref{eq:regatt} actively modifies $w_c$ and the student's parameters to minimize the KL-Divergence between the teacher's attention maps and the intermediate attention maps. 
At the same time, both $\mathcal{L}_\mathrm{CE}$ and $\mathcal{L}_\mathrm{KD}$ indirectly affect the adjustment of $w_c$, since $\mathbf{H}^S$ is related to the student model's predictions.
The hyperparameters $\lambda$ and $\mu$ serve as modulating factors to balance the weights of $\mathcal{L}_\mathrm{att}$ and $\mathcal{L}_\mathrm{KD}$. In line with Dynamic KD \cite{Li2021}, we modulate the value of $\lambda$ during training to adjust the supervision derived from $\mathcal{L}_\mathrm{att}$ and $\mathcal{L}_\mathrm{KD}$.
\begin{table*}[t]
\centering

\begin{tabular}{lccc}
\hline
Models           & De $\rightarrow$ En & De $\rightarrow$ Dsb & En $\rightarrow$ Zh \\ \hline
Teacher (6-layer)           & 36.79  $\scriptstyle \pm 0.51$                            & 38.68 $\scriptstyle \pm 2.02$                               & 23.97 $\scriptstyle \pm 0.34$                              \\ \hline 
\multicolumn{4}{c}{}\\ [-1.5ex]
\multicolumn{4}{c}{Student (3-layer)} \\ [0.5ex] \hline
No KD    & 36.24 $\scriptstyle \pm 0.50$                             & 35.88 $\scriptstyle \pm 1.97$                               & 23.20 $\scriptstyle \pm 0.31$                              \\
Vanilla KD~\cite{Hinton2015}               & 37.00 $\scriptstyle \pm 0.50$                             & 35.25 $\scriptstyle \pm 1.96$                              & 24.47 $\scriptstyle \pm 0.34$                              \\
Sequence KD~\cite{kim2016} & 36.61 $\scriptstyle \pm 0.51$ & 33.41 $\scriptstyle \pm 1.76$ & \textbf{24.67 $\scriptstyle \pm 0.35$} \\
Selective KD~\cite{Wang2021a}      & {\ul 37.30} $\scriptstyle \pm 0.50$                              & 35.94 $\scriptstyle \pm 1.95$                               & 22.66 $\scriptstyle \pm 0.32$                                                         \\
TinyBERT~\cite{Jiao2020}          & 37.24 $\scriptstyle\pm0.52$                            & 38.01 $\scriptstyle\pm1.80$                              & 24.31 $\scriptstyle\pm0.34$                           \\
MiniLM~\cite{Wang2020}            & 36.93 $\scriptstyle \pm 0.50$                              &  36.43 $\scriptstyle \pm 1.97 $                               & 24.32 $\scriptstyle \pm 0.33$                              \\
ALP-KD~\cite{Passban2021}            & 37.13 $\scriptstyle \pm 0.51$                              &  37.07 $\scriptstyle \pm 2.05 $                               & 24.36 $\scriptstyle \pm 0.33$                              \\

\textbf{A2D (4 heads)}        & 37.17 $\scriptstyle \pm 0.49$                              & {\ul 38.61} $\scriptstyle \pm 1.96$                               & {\ul 24.61} $\scriptstyle \pm 0.32$                              \\
\textbf{A2D (8 heads)}     & \textbf{37.75 $\scriptstyle \pm 0.51$}                             &      \textbf{39.49 $\scriptstyle \pm 2.06$}                        &        24.32 $\scriptstyle\pm 0.33$                    \\ \hline
\multicolumn{4}{c}{}\\ [-1.5ex]
\multicolumn{4}{c}{Student (2-layer)} \\ [0.5ex] \hline
No KD  & 35.56 $\scriptstyle \pm 0.51$                             &  {\ul 35.89} $\scriptstyle \pm 1.88$                               & \textbf{23.69 $\scriptstyle \pm 0.33$}                              \\
CKD-sc~\cite{Wu2020} & {\ul 35.73} $\scriptstyle \pm 0.49$                             & 32.90 $\scriptstyle \pm 1.85$                               & 22.55 $\scriptstyle \pm 0.33$                              \\
CKD-cc~\cite{Wu2020} & 35.46 $\scriptstyle \pm 0.48$                              & 31.28 $\scriptstyle \pm 1.74$                               & 22.86 $\scriptstyle \pm 0.31$                             \\
\textbf{A2D (4 heads)}    & \textbf{36.68 $\scriptstyle \pm 0.51$}                             & \textbf{37.06 $\scriptstyle \pm 1.94$}                               & {\ul 23.64} $\scriptstyle \pm 0.34$                              \\ \hline
\end{tabular}
\caption{BLEU scores of various KD approaches across language pairs. `No KD' denotes student models trained exclusively with the cross-entropy loss. Each baseline model incorporates 4 attention heads per layer. For a direct comparison with CKD, we train our students with A2D across 2 layers. The highest scores among student models are highlighted in bold, while the second highest are underlined. 
}
\label{tab:main}
\end{table*}
\begin{table*}[ht]
\centering
\begin{tabular}{c|cccccc}
\hline
Teacher     & No KD       & PKD    & CKD-rc      & CKD-oc   & \textbf{A2D}          & A2D (w/o $\mathcal{L}_\mathrm{KD}$) \\ \hline
27.70 $\scriptstyle \pm 0.65$ & 25.74 $\scriptstyle\pm 0.61$ & {\ul 23.38} & {\ul 24.14} & {\ul 23.97} & \textbf{26.37} $\scriptstyle \pm 0.64$ & 25.97 $\scriptstyle \pm 0.61$      \\ \hline
\end{tabular}
\caption{BLEU scores for WMT-2014 En $\xrightarrow{}$ De. `PKD' and `CKD' refer to Patient KD~\cite{Sun2019} and Combinatorial KD~\cite{Wu2020}, respectively. Underlined results are imported from CKD~\cite{Wu2020}. Our reproduced teacher and No KD student model, used for A2D training on En $\xrightarrow{}$ De, yielded slightly better BLEU scores than those reported in the CKD paper, which are 27.03 and 24.31 respectively. Nevertheless, the trend of their model underperforming compared to No KD remains consistent.}
\label{tab:wmt14ende}
\end{table*}
\subsection{Comparison with Previous Methods}
Previous works have also utilized attention distributions as knowledge features \cite{Jiao2020, Wang2020}. 
However, their approaches treat attention heads within the same layer as a singular unit for distillation.
Considering that the attention map from each head captures distinct information across the layer~\cite{voita-etal-2019-analyzing, gong2021pay}, establishing a rigid mapping between student and teacher layers imposes a potential loss of knowledge from the teacher.

In contrast, A2D facilitates a flexible alignment between each individual attention head of the teacher and student models, eliminating the need for pre-defined mapping combinations~\cite{Wu2020} or bucket divisions~\cite{Passban2021}. 
Additionally, A2D is not bound by architectural constraints, such as matching the number of heads or layers, or embedding dimensions, between the teacher and student models.
\section{Experiments}
\subsection{Datasets}
We use the public IWSLT and WMT datasets to evaluate our method on translation. 
The datasets of low-resource scenario include the IWSLT-2014 German $\xrightarrow{}$ English (De $\xrightarrow{}$ En), IWSLT-2017 English $\xrightarrow{}$ Chinese (En $\xrightarrow{}$ Zh), WMT-2022 German $\xrightarrow{}$ Lower Sorbian (De $\xrightarrow{}$ Dsb) translation. 
Tokenization is done with Subword-NMT \cite{Sennrich2016} for IWSLT-2014 dataset and Sentencepiece \cite{Kudo2018} for the others.
To prove the effectiveness of our method on high-resource scenarios, we evaluate NMT models on the WMT-2014 English  $\xrightarrow{}$ German (En $\xrightarrow{}$ De) datasets. we use newstest2013 datasets as a validation set and newstest2014 as the test set.
Data preparation for WMT-2014 En$\xrightarrow{}$De follows \cite{Vaswani2017} to ensure a fair comparison of baselines in Table \ref{tab:wmt14ende}.
\subsection{Distillation Settings}
For every experiment, the teacher and student are trained and evaluated with the same datasets.

\noindent\textbf{Low-resource translation} Teacher models are 6-layer Transformers \cite{Vaswani2017} with 4 attention heads, hidden dimensions, and the feed-forward dimension of 512, and 1024 for each.
Unless specified otherwise, student models are 3-layer Transformers with the same hyperparameters as the teacher, except for the number of layers.
For a fair comparison with CKD \cite{Wu2020} which reported results with 2-layer student networks, we additionally test our approach for 2-layer Transformer students. To show that A2D does not require the same number of attention heads between student and teacher, we also present the results with student models having 8 attention heads. For loss scaling, in Equation~\eqref{eq:attnloss}, $\lambda$ and $\mu$ are initially set to 1, and we set exponential decay on $\lambda$ at a rate of 0.9 over epochs. 

\noindent\textbf{High-resource translation} To properly scale the model to the high-resource data, we enlarge the teacher model and student model to 12-layer and 4-layer, respectively. 
The other hyperparameter settings such as attention heads and hidden dimension are set identically with \cite{Vaswani2017}.  
To adjust to the increased number of attention maps, We used $\lambda$ of $\mathcal{L}_\mathrm{att}$ as 0.1. 

\subsection{Baselines}
Selective KD~\cite{Wang2021a} is a variant of vanilla KD \cite{Hinton2015} which selectively chooses words to distill based on entropy. 
For selective KD, we use a ``word rate" of 0.5.
For TinyBERT~\cite{Jiao2020}, every other layer of the teacher model is correspondingly mapped to a layer in the student model.
MiniLM is originally proposed for knowledge distillation in the pre-training stage, so we augment the MiniLM objective with $\mathcal{L}_\mathrm{CE}$ and $\mathcal{L}_\mathrm{KD}$ for a fair comparison.
For ALP-KD~\cite{Passban2021}, we train the student with an attention mask spanned over all teacher layers.
For Combinatorial KD~\cite{Wu2020}~(CKD), `-sc', `-cc', `-rc', and `-oc' refer to its different layer mapping configurations.

\subsection{Results on Low-resource datasets}
Low-resource datasets present a unique challenge for NMT models, emphasizing the importance of effective knowledge distillation. 
Table~\ref{tab:main} summarizes all the results for low-resource NMT. 
Models are compared with BLEU scores computed using sacreBLEU~\cite{Matt2018}, with a confidence interval of 95\%, and the number of bootstrap resamples is 1000.
Previous KD methods, such as TinyBERT and MiniLM, impose a constraint that the number of heads in the student must match that of the teacher.
In contrast, A2D does not have this limitation regarding the number of attention heads, allowing us to train students with 8 heads using teachers with only 4 heads.
We also present the results of students with 8 heads in section~\ref{analysis}.
Doubling the number of attention heads from 4 to 8 does not alter the overall parameter count in the 'A2D (8heads)' model. This is achieved by proportionally reducing the dimension per head ($d_{head} = d_{model} / n_{head}$) while keeping the feature dimension ($d_{model}$) constant, thus ensuring a fair comparison.

In the De → En and De → Dsb language pairs, students trained with A2D yield higher BLEU scores than those trained with other KD methods. For the En → Zh language pair, A2D yields superior results compared to other feature-based KD baselines and is on par with Sequence-level KD.
In all language pair settings, students trained with A2D surpass the performance of their teachers despite having only half the number of transformer layers. 
Remarkably, the effectiveness of our method was most evident in the De $\xrightarrow[]{}$ Dsb dataset, which had the least amount of training data at 39K samples.
These results indicate that training with A2D allows students to achieve better generalization, particularly with low-resource training data.

To compare with CKD, we trained 2-layer students using A2D.  
While the score of CKD varies on its mapping option, our model consistently outperforms CKD regardless of the dataset or their mapping option.
Moreover, a 2-layer student in De $\xrightarrow{}$ En achieves results comparable to its 6-layer teacher.
The results not only validate the robustness of A2D but also show that our learned alignment of attention heads is more effective for knowledge transfer than the heuristic mapping of CKD.

In low-resource settings, a student model could outperform the teacher under the guidance of the teacher model. This is due to the enhanced generalization provided by the regularization effect of distillation, as discussed in \cite{DBLP:conf/nips/MobahiFB20, DBLP:conf/cvpr/YuanTLWF20}. The teacher model, often being larger and more complex, captures rich feature representations (such as attention maps) of the data. Through distillation, the student model learns these representations, which might be more generalizable than those learned from the relatively small amount of raw data alone. 

\subsection{Results on High-resource dataset}
To assess the versatility of our method, we extended our experiments to the high-resource dataset.
Table~\ref{tab:wmt14ende} describes our results with different baselines. 
From our observations, traditional KD techniques like Patient KD and Combinatorial KD did not enhance the performance of the student models for high-resource data. Surprisingly, they even underperform compared to a student model trained without any KD techniques.

However, a noticeable distinction arises when training the student model with the A2D approach, demonstrating a performance level similar to that of its teacher, even when modeled with only one-third number of the transformer layers compared to its teacher.
This performance gain remains even without integrating our method with vanilla KD, underscoring its effectiveness.
\section{Analysis and Discussion}\label{analysis}

\begin{table}[t]
    \centering
    \begin{tabular}{c|cc|cc}
        \hline
        $n_\mathrm{head}$ & {A2D} & {No KD} & $\Delta$BLEU & $\mathcal{L}_\mathrm{att}$ \\
        \hline
        2 & 36.33 & 35.79 & 0.54 & 0.036 \\
        \hline
        4 & 37.17 & 36.24 & 0.93 & 0.024 \\
        \hline
        8 & \textbf{37.75} & \textbf{36.25} & 1.5 & 0.016 \\
        \hline
        16 & 37.19 & 35.78 & 1.41 & 0.013 \\
        \hline
    \end{tabular}
    \caption{BLEU score and our attention distillation loss ($\mathcal{L}_\mathrm{att}$) at convergence over a different number of heads ($n_\mathrm{head}$) in A2D and No KD. $\Delta$BLEU indicates BLEU score difference between A2D and No KD.}
    \label{tab:nhead_trend}
\end{table}

\subsection{Effect of Fine-grained Alignment}
We focus on two attributes of A2D that enable fine-grained alignment: (1) dense, head-wise distillation (2) the use of an attention map as a feature.
Our hypothesis is that the detailed head-wise attention alignment is what gives A2D-trained students an edge in performance. In Table~\ref{tab:nhead_trend}, we present how our attention transfer loss and performance gap between students trained with A2D and students trained without KD vary with different numbers of attention heads at the point of convergence. Note that varying the number of heads in the student model does not change the total number of parameters.
Our observation suggests that as we reduce the $\mathcal{L}_\mathrm{att}$, $\Delta$BLEU becomes more pronounced, which indicates the effectiveness of our method. This inverse correlation between $\mathcal{L}_\mathrm{att}$ and $\Delta$BLEU as the number of heads increases could be attributed to our attention alignment module (AAM), which generates intermediate attention maps from original student maps.
AAM draws \textit{all} the student attention maps ($\mathbf{H}^S$)  to mimic \textit{each} teacher map ($\mathbf{H}^T_{c}$) via pointwise convolution operation; approximating $\mathbf{H}^T$ from an increased number of $\mathbf{H}^S$ is more feasible.
However, employing an excessive number of heads might degrade the performance by reducing the expressiveness of each student's attention head. 
We hypothesize that the choice of $n_{head}=16$ was overly complex for a model with specifications $d_{model}=512$ and reduced the capacity of each head.
This led to the model's optimal performance in BLEU with $n_{head}=8$, but recorded a dip in performance when escalated to $n_{head}=16$.

\begin{figure}[t]
\centering
\includegraphics[width=1.0\linewidth]{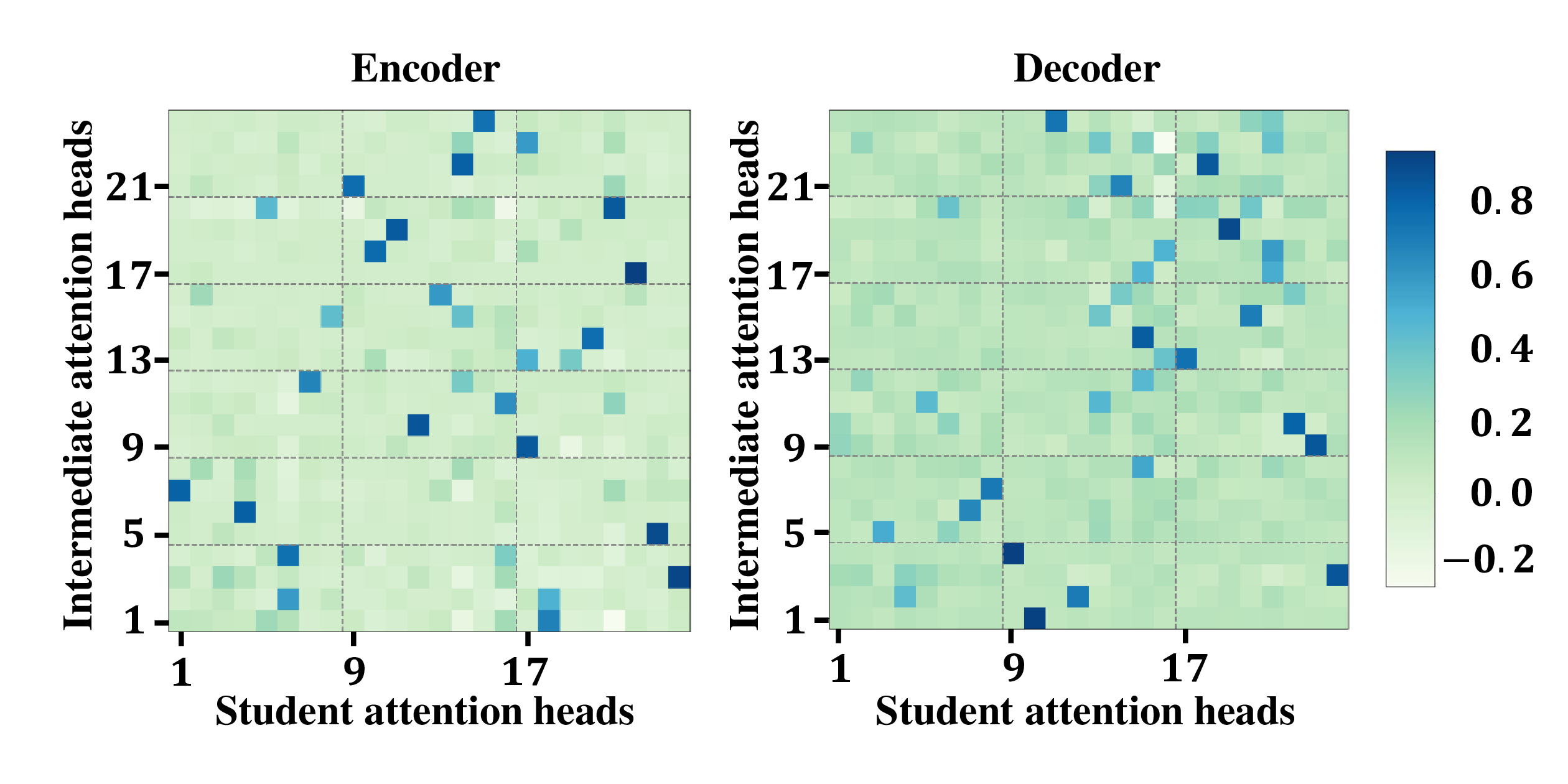}
\caption{Attention head connection weights in the trained AAM. Axes indicate attention head numbers in the student (3 layers of 8 heads) and teacher (6 layers of 4 heads) models. The dashed grid shows layer boundaries. Darker colors signify stronger connections. Best viewed in color.}
\label{fig:analysis}
\end{figure}

In Figure~\ref{fig:analysis}, we present the connectivity between attention maps learned by AAM (pointwise convolution layer) with heatmap. 
It shows that heads from students associate not only within but across the layers to form intermediate maps ($\mathbf{H}^I$) that are purposed to emulate the teacher maps ($\mathbf{H}^T$). 
This suggests that transferring knowledge using entire layers as units may not be the most effective approach for knowledge transfer.
To justify the claim, we also carried out evaluations on the layer-wise variant of A2D, as demonstrated in Table \ref{tab:layerwise}. 
Layer-wise A2D, which uses per-layer averaged maps as a knowledge feature, underperforms the original head-wise A2D by a significant margin on every language pair. 
This observation reinforces our hypothesis emphasizing the impact of our head-wise comparison approach in distillation.

\begin{table}[t]
\resizebox{\columnwidth}{!}{
\begin{tabular}{lcc}
\hline
Datasets & Layer-wise A2D   & \multicolumn{1}{c}{\begin{tabular}[c]{@{}c@{}}Head-wise A2D\\ (Default)\end{tabular}} \\ \hline
De $\xrightarrow{}$ En    & 37.01 $\scriptstyle\pm 0.49 $ & \textbf{37.75 $\scriptstyle\pm 0.49$}                                                                    \\
De $\xrightarrow{}$ Dsb   &   36.70 $\scriptstyle\pm 1.96 $               &   \textbf{39.49 $\scriptstyle \pm 2.06$}                                                                                       \\
En $\xrightarrow{}$ Zh    &    24.07 $\scriptstyle\pm 0.34 $              &         \textbf{24.32 $\scriptstyle \pm 0.33$}                                                                               \\ \hline
\end{tabular}
}
\caption{Comparison of our original A2D (head-wise distillation) with its layer-wise counterpart.} 
\label{tab:layerwise}
\end{table}
\begin{table*}[t]
\centering
\begin{tabular}{lccccc}
\hline
Datasets & A2D (All) & A2D (Enc-Self) & A2D (Dec-All) & A2D (Dec-Self) & A2D (Dec-Cross) \\ \hline
De $\xrightarrow{}$ En & \textbf{37.75 $\scriptstyle \pm 0.51$} & 37.39 $\scriptstyle \pm 0.50$ & 37.21 $\scriptstyle \pm 0.51$ & 37.16 $\scriptstyle \pm 0.49$ & 37.25 $\scriptstyle \pm 0.51$ \\
De $\xrightarrow{}$ Dsb & \textbf{39.49 $\scriptstyle \pm 2.06$} & 36.80 $\scriptstyle \pm 2.01$ & 37.81 $\scriptstyle \pm 1.95$ & 36.81 $\scriptstyle \pm 2.13$ & 37.66 $\scriptstyle \pm 1.97$ \\
En $\xrightarrow{}$ Zh & 24.32 $\scriptstyle \pm 0.33$ & \textbf{24.77 $\scriptstyle \pm 0.32$} & 24.75 $\scriptstyle \pm 0.35$ & 24.67 $\scriptstyle \pm 0.33$ & 24.64 $\scriptstyle \pm0.35 $ \\ \hline
\end{tabular}
\caption{Ablation studies by applying A2D on different parts of Transformer. Dec-All setting indicates that both self-attention and cross-attention maps are used for A2D. For Dec-Self and Dec-Cross settings, we increased the weight of $L_\mathrm{att}^\mathrm{dec-self}$ and $L_\mathrm{att}^\mathrm{dec-cross}$ from $1/2$ to 1 respectively to match the loss scale on the Dec-All setting.}
\label{tab:decoder_appendix}
\end{table*}

\begin{table*}[t]
    \centering
    \small
    \begin{tabular}{l|c|c|c|c|c|c|c|c|c|c}
        \toprule
        Model & \#Params & CoLA & MNLI-(m/mm) & SST-2 & QNLI & MRPC & QQP & RTE & STS-B & Avg \\
                           &       & (Mcc)  & (Acc)       & (Acc) & (Acc) & (F1) & (Acc) & (Acc) & (Spear) &  \\
        \midrule
        $\text{BERT}_{base}$ & 110M & 58.7 & 84.5/84.5 & 91.7 & 91.3 & 89.0 & 91.1 & 67.9 & 89.5 & 82.9 \\
        $\text{BERT}_{6}$ & 66M & 51.2 & 81.7/82.6 & 91.0 & 89.3 & 89.2 & 90.4 & 66.1 & 88.3 & 80.9 \\ \hline
        PD & 66M & - & 82.5/83.4 & 91.1 & 89.4 & \textbf{89.4} & {\ul 90.7} & 66.7 & - & - \\
        PKD & 66M & 45.5 & 81.3/- & 91.3 & 88.4 & 85.7 & 88.4 & 66.5 & 86.2 & 79.2 \\
        TinyBERT & 66M & {\ul 53.8} & {\ul 83.1/83.4} & \textbf{92.3} & {\ul 89.9} & 88.8 & 90.5 & {\ul 66.9} & {\ul 88.3} & {\ul 81.7} \\
        \textbf{A2D} & 66M & \textbf{58.8} & \textbf{83.2/83.5} & {\ul 91.7} & \textbf{90.3} & {\ul 89.2} & \textbf{90.9} & \textbf{67.5} & \textbf{88.7} &  \textbf{82.5}\\
        \bottomrule
    \end{tabular}
    \caption{Evaluation results on the dev set of GLUE Benchmark. We use $\text{BERT}_{base}$ and $\text{BERT}_{6}$ as teacher and student model, respectively.  Both $\text{BERT}_{6}$, a 6-layer smaller variant of BERT, and `PD', a distilled model, are released by \citeauthor{Turc2019WellReadSL}. The results of baselines are imported from ~\citeauthor{park-etal-2021-distilling}.}
    \label{tab:glue}
\end{table*}

\subsection{Decoder Distillation}
While previous feature-based KD methods focused on distilling encoder-only models~\cite{Jiao2020, Wang2020, Passban2021} and did not discover effective KD settings for the decoder in NMT tasks~\cite{Wu2020}, A2D distills both the encoder and decoder.

Our claim regarding the effectiveness of our method on the decoder is supported by the ablation studies presented in Table~\ref{tab:decoder_appendix}. Generally, applying A2D to both the encoder and decoder together yielded the best results. In the En $\xrightarrow{}$ Zh direction, there is a slight performance degradation when using integrated encoder and decoder distillation compared to using encoder-only or decoder-only distillation. Nevertheless, our model trained with decoder-only distillation outperforms the one with distillation on both the encoder and decoder, demonstrating its effectiveness on the decoder.

To investigate why A2D is effective on the decoder, we examine the AAM of the decoder, as described in Figure~\ref{fig:analysis}.
From the heatmap, we observe that the connection between heads tends to be sparse at the encoder level, with most values near 0.
Conversely, in the decoder, the connection between heads is more evenly distributed, with values shifting away from 0 and closer to 0.2.
Based on this observation, we believe that our soft, fine-grained connections between teacher-student features led to more successful KD for decoder tasks, as compared to the strict on/off connections proposed in earlier KD studies.

\subsection{Effectiveness on Different Tasks}
In this study, we assessed our method using NMT to demonstrate its potential beyond tasks solely associated with encoders.
Additionally, we present performance metrics for the A2D method when applied to encoder-only models on natural language understanding benchmarks given that most of the KD in natural language processing research has focused on the encoder-only models and tasks. 
Specifically, we applied our method to BERT~\cite{Devlin2019} distillation, benchmarked on the GLUE~\cite{wang-etal-2018-glue} dataset. The comparative results with baselines are presented in Table~\ref{tab:glue}. For training, we used fine-tuned teacher models for each GLUE task and then applied A2D to the corresponding student model. We experimented with the hyperparameter $\lambda$ of Equation~\eqref{eq:ceatt}, selecting from the values \{0.01, 0.02, 0.05, 0.1\}. For all other configurations, we followed BERT's settings. Notably, our method demonstrated superior performance over encoder-oriented baselines, even without dedicated task-specific hyperparameter tuning.
\section{Conclusion}
In this paper, we introduce Align-to-Distill, a novel approach to knowledge distillation that enables a detailed alignment of attention heads between teacher and student models. We propose a strategy to overcome the need for heuristic feature mapping in a learnable manner. Our approach shows promising results in decoder distillation, effectively compressing models while preserving translation quality.
\section{Limitations and Future work}
Although A2D is architecturally flexible without constraints on hidden size or number of attention heads, the teacher and student models must share the same vocabulary. This requirement potentially restricts its broader applicability. Also, while our work demonstrates A2D's effectiveness on the decoder module in Table~\ref{tab:decoder_appendix}, we have not yet tested A2D on decoder-only models. The scope of this paper primarily focuses on encoder-decoder-based translation models. In future work, we plan to extend A2D's application to decoder-only models. Moreover, the concept of A2D may encompass a broader range of architectures in future work, as the idea of adaptively aligning features is not restricted to using attention as a feature. 

\bibliography{acl_latex}

\begin{thebibliography}{37}
\expandafter\ifx\csname natexlab\endcsname\relax\def\natexlab#1{#1}\fi

\bibitem[{Ba and Caruana(2014)}]{Ba2014}
Jimmy Ba and Rich Caruana. 2014.
\newblock Do deep nets really need to be deep?
\newblock In \emph{Advances in Neural Information Processing Systems: Annual Conference on Neural Information Processing Systems, {(NIPS)}}, pages 2654--2662.

\bibitem[{Bahdanau et~al.(2015)Bahdanau, Cho, and Bengio}]{Bahdanau2015}
Dzmitry Bahdanau, Kyunghyun Cho, and Yoshua Bengio. 2015.
\newblock Neural machine translation by jointly learning to align and translate.
\newblock In \emph{International Conference on Learning Representations, {(ICLR)} Conference Track Proceedings}.

\bibitem[{Bucila et~al.(2006)Bucila, Caruana, and Niculescu{-}Mizil}]{Bucila2006}
Cristian Bucila, Rich Caruana, and Alexandru Niculescu{-}Mizil. 2006.
\newblock Model compression.
\newblock In \emph{Proceedings of International Conference on Knowledge Discovery and Data Mining, {(SIGKDD)}}, pages 535--541.

\bibitem[{Devlin et~al.(2019)Devlin, Chang, Lee, and Toutanova}]{Devlin2019}
Jacob Devlin, Ming{-}Wei Chang, Kenton Lee, and Kristina Toutanova. 2019.
\newblock {BERT:} pre-training of deep bidirectional transformers for language understanding.
\newblock In \emph{Proceedings of the Conference of the North American Chapter of the Association for Computational Linguistics: Human Language Technologies, {(NAACL-HLT)}}, pages 4171--4186.

\bibitem[{Gong et~al.(2021)Gong, Tang, Pino, and Li}]{gong2021pay}
Hongyu Gong, Yun Tang, Juan Pino, and Xian Li. 2021.
\newblock Pay better attention to attention: Head selection in multilingual and multi-domain sequence modeling.
\newblock \emph{Advances in Neural Information Processing Systems}, 34:2668--2681.

\bibitem[{Gou et~al.(2021)Gou, Yu, Maybank, and Tao}]{DBLP:journals/ijcv/GouYMT21}
Jianping Gou, Baosheng Yu, Stephen~J. Maybank, and Dacheng Tao. 2021.
\newblock Knowledge distillation: {A} survey.
\newblock \emph{Int. J. Comput. Vis.}, 129(6):1789--1819.

\bibitem[{Gu et~al.(2017)Gu, Neubig, Cho, and Li}]{gu2017}
Jiatao Gu, Graham Neubig, Kyunghyun Cho, and Victor~O.K. Li. 2017.
\newblock Learning to translate in real-time with neural machine translation.
\newblock In \emph{Proceedings of the 15th Conference of the {E}uropean Chapter of the Association for Computational Linguistics: Volume 1, Long Papers}, pages 1053--1062. Association for Computational Linguistics.

\bibitem[{Hinton et~al.(2015)Hinton, Vinyals, and Dean}]{Hinton2015}
Geoffrey~E. Hinton, Oriol Vinyals, and Jeffrey Dean. 2015.
\newblock Distilling the knowledge in a neural network.
\newblock \emph{CoRR}, abs/1503.02531.

\bibitem[{Jiao et~al.(2020)Jiao, Yin, Shang, Jiang, Chen, Li, Wang, and Liu}]{Jiao2020}
Xiaoqi Jiao, Yichun Yin, Lifeng Shang, Xin Jiang, Xiao Chen, Linlin Li, Fang Wang, and Qun Liu. 2020.
\newblock Tinybert: Distilling {BERT} for natural language understanding.
\newblock In \emph{Findings of the Association for Computational Linguistics: {(EMNLP)}}, pages 4163--4174.

\bibitem[{Joyce(2011)}]{KLD}
James~M Joyce. 2011.
\newblock Kullback-leibler divergence.
\newblock In \emph{International encyclopedia of statistical science}, pages 720--722. Springer.

\bibitem[{Kasai et~al.(2020)Kasai, Pappas, Peng, Cross, and Smith}]{Kasai2020DeepES}
Jungo Kasai, Nikolaos Pappas, Hao Peng, James Cross, and Noah~A. Smith. 2020.
\newblock Deep encoder, shallow decoder: Reevaluating non-autoregressive machine translation.
\newblock In \emph{International Conference on Learning Representations}.

\bibitem[{Kim and Rush(2016)}]{kim2016}
Yoon Kim and Alexander~M. Rush. 2016.
\newblock Sequence-level knowledge distillation.
\newblock In \emph{Proceedings of the 2016 Conference on Empirical Methods in Natural Language Processing}, pages 1317--1327. Association for Computational Linguistics.

\bibitem[{Kudo and Richardson(2018)}]{Kudo2018}
Taku Kudo and John Richardson. 2018.
\newblock Sentencepiece: {A} simple and language independent subword tokenizer and detokenizer for neural text processing.
\newblock In \emph{Proceedings of the Conference on Empirical Methods in Natural Language Processing, {(EMNLP)}}, pages 66--71.

\bibitem[{Lewis et~al.(2020)Lewis, Liu, Goyal, Ghazvininejad, Mohamed, Levy, Stoyanov, and Zettlemoyer}]{Lewis2020}
Mike Lewis, Yinhan Liu, Naman Goyal, Marjan Ghazvininejad, Abdelrahman Mohamed, Omer Levy, Veselin Stoyanov, and Luke Zettlemoyer. 2020.
\newblock {BART:} denoising sequence-to-sequence pre-training for natural language generation, translation, and comprehension.
\newblock In \emph{Proceedings of the 58th Annual Meeting of the Association for Computational Linguistics, {(ACL)}}, pages 7871--7880.

\bibitem[{Li et~al.(2021)Li, Lin, Ren, Li, Zhou, and Sun}]{Li2021}
Lei Li, Yankai Lin, Shuhuai Ren, Peng Li, Jie Zhou, and Xu~Sun. 2021.
\newblock Dynamic knowledge distillation for pre-trained language models.
\newblock In \emph{Proceedings of the Conference on Empirical Methods in Natural Language Processing, {{EMNLP}}}, pages 379--389.

\bibitem[{Liu et~al.(2019)Liu, Ott, Goyal, Du, Joshi, Chen, Levy, Lewis, Zettlemoyer, and Stoyanov}]{Liu2019}
Yinhan Liu, Myle Ott, Naman Goyal, Jingfei Du, Mandar Joshi, Danqi Chen, Omer Levy, Mike Lewis, Luke Zettlemoyer, and Veselin Stoyanov. 2019.
\newblock Roberta: {A} robustly optimized {BERT} pretraining approach.
\newblock \emph{CoRR}, abs/1907.11692.

\bibitem[{Mobahi et~al.(2020)Mobahi, Farajtabar, and Bartlett}]{DBLP:conf/nips/MobahiFB20}
Hossein Mobahi, Mehrdad Farajtabar, and Peter~L. Bartlett. 2020.
\newblock Self-distillation amplifies regularization in hilbert space.
\newblock In \emph{Advances in Neural Information Processing Systems 33: Annual Conference on Neural Information Processing Systems 2020, NeurIPS 2020, December 6-12, 2020, virtual}.

\bibitem[{Park et~al.(2021)Park, Kim, and Yang}]{park-etal-2021-distilling}
Geondo Park, Gyeongman Kim, and Eunho Yang. 2021.
\newblock Distilling linguistic context for language model compression.
\newblock In \emph{Proceedings of the 2021 Conference on Empirical Methods in Natural Language Processing}, pages 364--378. Association for Computational Linguistics.

\bibitem[{Passban et~al.(2021)Passban, Wu, Rezagholizadeh, and Liu}]{Passban2021}
Peyman Passban, Yimeng Wu, Mehdi Rezagholizadeh, and Qun Liu. 2021.
\newblock {ALP-KD:} attention-based layer projection for knowledge distillation.
\newblock In \emph{Conference on Innovative Applications of Artificial Intelligence, {(IAAI)}, The Symposium on Educational Advances in Artificial Intelligence, {(EAAI)}}, pages 13657--13665.

\bibitem[{Post(2018)}]{Matt2018}
Matt Post. 2018.
\newblock A call for clarity in reporting {BLEU} scores.
\newblock In \emph{Proceedings of the Third Conference on Machine Translation: Research Papers}, pages 186--191. Association for Computational Linguistics.

\bibitem[{Romero et~al.(2015)Romero, Ballas, Kahou, Chassang, Gatta, and Bengio}]{Romero2015}
Adriana Romero, Nicolas Ballas, Samira~Ebrahimi Kahou, Antoine Chassang, Carlo Gatta, and Yoshua Bengio. 2015.
\newblock Fitnets: Hints for thin deep nets.
\newblock In \emph{International Conference on Learning Representations, {(ICLR)}}.

\bibitem[{Sanh et~al.(2019)Sanh, Debut, Chaumond, and Wolf}]{Sanh2019}
Victor Sanh, Lysandre Debut, Julien Chaumond, and Thomas Wolf. 2019.
\newblock Distilbert, a distilled version of {BERT:} smaller, faster, cheaper and lighter.
\newblock \emph{CoRR}, abs/1910.01108.

\bibitem[{Sennrich et~al.(2016)Sennrich, Haddow, and Birch}]{Sennrich2016}
Rico Sennrich, Barry Haddow, and Alexandra Birch. 2016.
\newblock Neural machine translation of rare words with subword units.
\newblock In \emph{Proceedings of the Annual Meeting of the Association for Computational Linguistics, {(ACL)}}.

\bibitem[{Shleifer and Rush(2020)}]{Shleifer2020PreSum}
Sam Shleifer and Alexander~M. Rush. 2020.
\newblock Pre-trained summarization distillation.
\newblock \emph{CoRR}, abs/2010.13002.

\bibitem[{Sun et~al.(2019)Sun, Cheng, Gan, and Liu}]{Sun2019}
Siqi Sun, Yu~Cheng, Zhe Gan, and Jingjing Liu. 2019.
\newblock Patient knowledge distillation for {BERT} model compression.
\newblock In \emph{Proceedings of the Conference on Empirical Methods in Natural Language Processing and the International Joint Conference on Natural Language Processing, {(EMNLP-IJCNLP)}}, pages 4322--4331.

\bibitem[{Sun et~al.(2020)Sun, Yu, Song, Liu, Yang, and Zhou}]{Sun2020}
Zhiqing Sun, Hongkun Yu, Xiaodan Song, Renjie Liu, Yiming Yang, and Denny Zhou. 2020.
\newblock Mobilebert: a compact task-agnostic {BERT} for resource-limited devices.
\newblock In \emph{Proceedings of the Annual Meeting of the Association for Computational Linguistics, {(ACL)}}, pages 2158--2170.

\bibitem[{Sutskever et~al.(2014)Sutskever, Vinyals, and Le}]{Sutskever2014}
Ilya Sutskever, Oriol Vinyals, and Quoc~V. Le. 2014.
\newblock Sequence to sequence learning with neural networks.
\newblock In \emph{Advances in Neural Information Processing Systems: Annual Conference on Neural Information Processing Systems, {(NIPS)}}, pages 3104--3112.

\bibitem[{Turc et~al.(2019)Turc, Chang, Lee, and Toutanova}]{Turc2019WellReadSL}
Iulia Turc, Ming-Wei Chang, Kenton Lee, and Kristina Toutanova. 2019.
\newblock Well-read students learn better: On the importance of pre-training compact models.
\newblock \emph{arXiv: Computation and Language}.

\bibitem[{Vaswani et~al.(2017)Vaswani, Shazeer, Parmar, Uszkoreit, Jones, Gomez, Kaiser, and Polosukhin}]{Vaswani2017}
Ashish Vaswani, Noam Shazeer, Niki Parmar, Jakob Uszkoreit, Llion Jones, Aidan~N. Gomez, Lukasz Kaiser, and Illia Polosukhin. 2017.
\newblock Attention is all you need.
\newblock In \emph{Advances in Neural Information Processing Systems: Annual Conference on Neural Information Processing Systems, {(NIPS)}}, pages 5998--6008.

\bibitem[{Voita et~al.(2019)Voita, Talbot, Moiseev, Sennrich, and Titov}]{voita-etal-2019-analyzing}
Elena Voita, David Talbot, Fedor Moiseev, Rico Sennrich, and Ivan Titov. 2019.
\newblock Analyzing multi-head self-attention: Specialized heads do the heavy lifting, the rest can be pruned.
\newblock In \emph{Proceedings of the 57th Annual Meeting of the Association for Computational Linguistics}, pages 5797--5808. Association for Computational Linguistics.

\bibitem[{Wang et~al.(2018)Wang, Singh, Michael, Hill, Levy, and Bowman}]{wang-etal-2018-glue}
Alex Wang, Amanpreet Singh, Julian Michael, Felix Hill, Omer Levy, and Samuel Bowman. 2018.
\newblock {GLUE}: A multi-task benchmark and analysis platform for natural language understanding.
\newblock In \emph{Proceedings of the 2018 {EMNLP} Workshop {B}lackbox{NLP}: Analyzing and Interpreting Neural Networks for {NLP}}, pages 353--355. Association for Computational Linguistics.

\bibitem[{Wang et~al.(2021)Wang, Yan, Meng, and Zhou}]{Wang2021a}
Fusheng Wang, Jianhao Yan, Fandong Meng, and Jie Zhou. 2021.
\newblock Selective knowledge distillation for neural machine translation.
\newblock In \emph{Proceedings of the Annual Meeting of the Association for Computational Linguistics and the International Joint Conference on Natural Language Processing, {(ACL/IJCNLP)}}, pages 6456--6466.

\bibitem[{Wang et~al.(2020)Wang, Wei, Dong, Bao, Yang, and Zhou}]{Wang2020}
Wenhui Wang, Furu Wei, Li~Dong, Hangbo Bao, Nan Yang, and Ming Zhou. 2020.
\newblock Minilm: Deep self-attention distillation for task-agnostic compression of pre-trained transformers.
\newblock In \emph{Advances in Neural Information Processing Systems: Annual Conference on Neural Information Processing Systems, {(NeurIPS)}}.

\bibitem[{Wu et~al.(2020)Wu, Passban, Rezagholizadeh, and Liu}]{Wu2020}
Yimeng Wu, Peyman Passban, Mehdi Rezagholizadeh, and Qun Liu. 2020.
\newblock Why skip if you can combine: {A} simple knowledge distillation technique for intermediate layers.
\newblock In \emph{Proceedings of the Conference on Empirical Methods in Natural Language Processing, {(EMNLP)}}, pages 1016--1021.

\bibitem[{Yuan et~al.(2020)Yuan, Tay, Li, Wang, and Feng}]{DBLP:conf/cvpr/YuanTLWF20}
Li~Yuan, Francis E.~H. Tay, Guilin Li, Tao Wang, and Jiashi Feng. 2020.
\newblock Revisiting knowledge distillation via label smoothing regularization.
\newblock In \emph{2020 {IEEE/CVF} Conference on Computer Vision and Pattern Recognition, {CVPR} 2020, Seattle, WA, USA, June 13-19, 2020}, pages 3902--3910. Computer Vision Foundation / {IEEE}.

\bibitem[{Zagoruyko and Komodakis(2017)}]{Zagoruyko2017}
Sergey Zagoruyko and Nikos Komodakis. 2017.
\newblock Paying more attention to attention: Improving the performance of convolutional neural networks via attention transfer.
\newblock In \emph{International Conference on Learning Representations, {(ICLR)}}.

\bibitem[{Zhou et~al.(2022)Zhou, Eisner, Newman, Platanios, and Thomson}]{zhou2022}
Jiawei Zhou, Jason Eisner, Michael Newman, Emmanouil~Antonios Platanios, and Sam Thomson. 2022.
\newblock Online semantic parsing for latency reduction in task-oriented dialogue.
\newblock In \emph{Proceedings of the 60th Annual Meeting of the Association for Computational Linguistics (Volume 1: Long Papers)}, pages 1554--1576. Association for Computational Linguistics.

\end{thebibliography}

\end{document}